\documentclass[runningheads]{llncs}
\usepackage{graphicx}
\usepackage{amsmath,amssymb} % define this before the line numbering.
\usepackage{color}

\usepackage{microtype}
\usepackage[pagebackref,breaklinks,colorlinks]{hyperref}

\begin{document}
	%\pagestyle{headings}
	%\mainmatter
	
	%\def\ACCV22SubNumber{818}  % Insert your submission number here
	
	%===========================================================
	\title{Exploring Adversarially Robust Training for Unsupervised Domain Adaptation} % Replace with your title
	\titlerunning{Asian Conference on Computer Vision 2022}
	\authorrunning{S.-Y. Lo et al.}
	
	\author{Shao-Yuan Lo \and Vishal M. Patel}
	\institute{Dept. of Electrical and Computer Engineering, Johns Hopkins University \\
		\email{\{sylo, vpatel36\}@jhu.edu}}
	
	\maketitle
	
	%===========================================================
	\begin{abstract}
		Unsupervised Domain Adaptation (UDA) methods aim to transfer knowledge from a labeled source domain to an unlabeled target domain. UDA has been extensively studied in the computer vision literature. Deep networks have been shown to be vulnerable to adversarial attacks. However, very little focus is devoted to improving the adversarial robustness of deep UDA models, causing serious concerns about model reliability. Adversarial Training (AT) has been considered to be the most successful adversarial defense approach. Nevertheless, conventional AT requires ground-truth labels to generate adversarial examples and train models, which limits its effectiveness in the unlabeled target domain. In this paper, we aim to explore AT to robustify UDA models: How to enhance the unlabeled data robustness via AT while learning domain-invariant features for UDA? To answer this question, we provide a systematic study into multiple AT variants that can potentially be applied to UDA. Moreover, we propose a novel Adversarially Robust Training method for UDA accordingly, referred to as ARTUDA. Extensive experiments on multiple adversarial attacks and UDA benchmarks show that ARTUDA consistently improves the adversarial robustness of UDA models. Code is available at \url{https://github.com/shaoyuanlo/ARTUDA}
	\end{abstract}
	
	%===========================================================
	\section{Introduction} \label{sec:intro}
	Recent advances in image recognition have enjoyed remarkable success via deep supervised learning \cite{he2016deep,krizhevsky2012imagenet,Zagoruyko2016WRN}. However, the domain shift problem is very common in real-world applications, \textit{i.e.}, source and target domains have different data characteristics. Furthermore, it is costly and labor-intensive to collect the ground-truth labels of target data. To address this issue, Unsupervised Domain Adaptation (UDA) methods have been developed in which the objective is to transfer the knowledge from a labeled source dataset to an unlabeled target dataset. Most existing UDA approaches rely on minimizing distribution discrepancy between source and target domains to learn domain-invariant representations \cite{ganin2015unsupervised,ganin2016domain,long2015learning,long2017conditional,long2017deep,tzeng2017adversarial}. Although these approaches achieve impressive performance, they do not consider the robustness against adversarial attacks \cite{biggio2013evasion,Szegedy2014Intriguing}, which causes critical concerns. 
	
	Adversarial attacks pose serious security risks to deep networks. In other words, deep networks could suffer from dramatic performance degradation in the presence of carefully crafted perturbations. To defend against adversarial attacks, various defense mechanisms have been proposed \cite{goodfellow2015explaining,guo2017countering,kannan2018adversarial,lo2021error,madry2018towards,raff2019barrage,zhang2019theoretically}. Currently, Adversarial Training (AT) based defenses \cite{goodfellow2015explaining,kannan2018adversarial,madry2018towards,zhang2019theoretically} have been considered the most effective, especially under the white-box setting \cite{obfuscated}. The core idea is to train a model on adversarial examples that are generated on the fly according to the model’s current parameters. Nevertheless, conventional AT requires ground-truth labels to generate adversarial examples. This makes it not applicable to the UDA problem since UDA considers the scenario that label information is unavailable to a target domain. A nearly contemporary work \cite{awais2021adversarial} resorts to external adversarially pre-trained ImageNet models as teacher models to distill robustness knowledge. However, its performance is highly sensitive to the teacher models’ perturbation budget, architecture, \textit{etc.}, which limits the flexibility in a wide range of uses. Another very recent work \cite{yang2021exploring} uses an external pre-trained UDA model to produce pseudo labels for doing AT on target data. Unfortunately, we show that it suffers from suboptimal accuracy and robustness against white-box attacks.
	
	\begin{figure}[!t]
		\centering
		\includegraphics[width=0.96\textwidth]{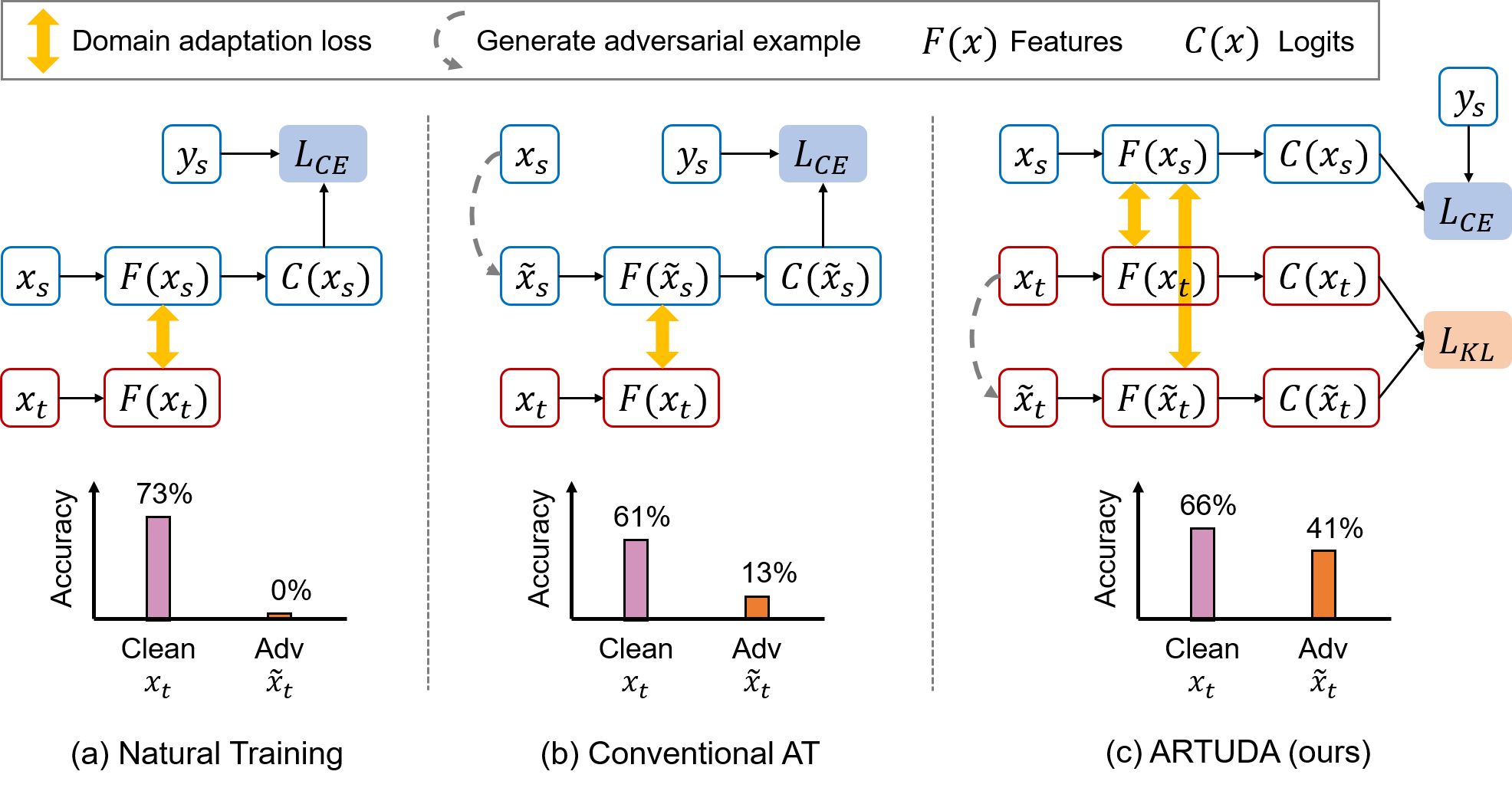}
		\caption{Overview of the proposed ARTUDA and its importance. $L_{CE}$: Cross-entropy loss. $L_{KL}$: KL divergence loss. Compared to conventional AT \cite{madry2018towards}, ARTUDA significantly improves adversarial robustness while maintaining decent clean accuracy. We use DANN \cite{ganin2016domain} with ResNet-50 \cite{he2016deep} backbone, the VisDA-2017 \cite{peng2017visda} dataset, and the PGD-20 \cite{madry2018towards} attack for this experiment.}
		\label{fig:main}
		%\vspace*{-\baselineskip}
	\end{figure}
	
	Given the above observations, intuitive questions emerge: \textit{Can we develop an AT algorithm specifically for the UDA problem? How to improve the unlabeled data robustness via AT while learning domain-invariant features for UDA?} In this paper, we seek to answer these questions by systematically studying multiple AT variants that can potentially be applied to UDA. First, we apply a conventional AT \cite{madry2018towards} to an UDA model to see its effectiveness. In other words, the AT is performed on only the labeled source data. Second, inspired by \cite{kannan2018adversarial,zhang2019theoretically}, we attempt to train models by minimizing the difference between the output logits of clean target data and the corresponding adversarial examples. With this, we can conduct a kind of AT directly on the target data in a self-supervised manner. We call it \textit{Self-Supervised Adversarial Training} or \textit{Self-Supervised AT}. Next, we look into the effects of clean images and adversarial examples in the AT for UDA. We present the trade-off behind different AT variants. Last, we observe that Batch Normalization (BN) \cite{ioffe2015batch} plays an important role in the AT for UDA. The feature statistic estimations at training time would affect an UDA model’s robustness.
	
	Through these investigations, we propose a novel Adversarially Robust Training method for UDA accordingly, referred to as ARTUDA. It uses both source and target data for training and does not require target domain labels, so it is feasible for UDA. Moreover, it does not need guidance from external models such as adversarially pre-trained models and pre-trained UDA models. Fig.~\ref{fig:main} illustrates an overview and the importance of the proposed ARTUDA. The naturally trained (\textit{i.e.}, train with only clean data) model’s accuracy decreases to 0\% under an adversarial attack. Conventional AT \cite{madry2018towards} improves robust accuracy to 13\% but sacrifices clean accuracy. As can be seen, ARTUDA significantly increases robust accuracy to 41\% while maintaining better clean accuracy. This shows that our method can improve unlabeled data robustness and learn domain-invariant features simultaneously for UDA. To the best of our knowledge, ARTUDA is the first AT-based UDA defense that is robust against white-box attacks. In Sec.~\ref{sec:expts}, we extensively evaluate ARTUDA on five adversarial attacks, three datasets and three different UDA algorithms. The results demonstrate its wide range of effectiveness.
	
	Our main contributions can be summarized as follows: (i) We provide a systematic study into various AT methods that are suitable for UDA. We believe that such experimental analysis would provide useful insight into this relatively unexplored research direction. (ii) We propose ARTUDA, a new AT method specifically designed for UDA. To the best of our knowledge, it is the first AT-based UDA defense method that is robust against white-box attacks. (iii) Comprehensive experiments show that ARTUDA consistently improves UDA models’ adversarial robustness under multiple attacks and datasets.
	
	%------------------------------------------------------------------------
	\section{Related Work} \label{sec:related}
	\noindent \textbf{Unsupervised domain adaptation.}
	UDA considers the scenario that a source dataset contains images with category labels, while label information is unavailable to a target dataset. Most popular approaches attempt to transfer knowledge from the labeled source domain to the unlabeled target domain \cite{ganin2015unsupervised,ganin2016domain,long2015learning,long2017conditional,long2017deep,tzeng2017adversarial}. DANN \cite{ganin2016domain} proposes to use a domain discriminator that distinguishes between source and target features, and the feature extractor is trained to fool it via GAN \cite{goodfellow2014generative} learning scheme. ADDA \cite{tzeng2017adversarial} combines DANN with discriminative feature learning. CDAN \cite{long2017conditional} extends DANN using a class-conditional adversarial game. JAN \cite{long2017deep} aligns the joint distributions of domain-specific layers between two domains. Nevertheless, these works do not take adversarial robustness into consideration.
	
	RFA \cite{awais2021adversarial} and ASSUDA \cite{yang2021exploring} are the most related works in the literature, which are nearly contemporary with our work. They are the first to focus on UDA’s adversarial robustness, but we would like to point out the clear differences from our work. RFA leverages external adversarially pre-trained ImageNet models as teacher models to distill robustness knowledge. Its performance is highly sensitive to the teacher models’ setup, such as perturbation budget, architecture and the number of teachers. AT on ImageNet is very expensive, so it is not always easy to obtain the preferred teacher models. In contrast, we propose a method that directly performs AT on a given UDA task, enjoying maximum flexibility. ASSUDA aims at semantic segmentation and considers only weak black-box attacks. It employs an external pre-trained UDA model to produce pseudo labels for target data, then uses the pseudo labels to do AT. However, we show that this approach has suboptimal accuracy and robustness against white-box attacks. In contrast, our method is robust under both black-box and white-box settings.
	
	\noindent \textbf{Adversarial attack and defense.}
	Previous studies reveal that deep networks are vulnerable to adversarial examples \cite{biggio2013evasion,Szegedy2014Intriguing}. Many adversarial attack algorithms have been proposed, such as Fast Gradient Sign Method (FGSM) \cite{goodfellow2015explaining}, Projected Gradient Descent (PGD) \cite{madry2018towards}, Momentum Iterative FGSM (MI-FGSM) \cite{dong2018boosting} and Multiplicative Adversarial Example (MultAdv) \cite{lo2021multav}.
	
	Various adversarial defense mechanisms have also been introduced, where AT-based defenses \cite{goodfellow2015explaining,kannan2018adversarial,madry2018towards,zhang2019theoretically} are considered the most effective, especially under the white-box setting \cite{obfuscated}. AT trains a model on adversarial examples that are generated on the fly according to the model’s current parameters. The most commonly used AT approaches include Madry’s AT scheme (PGD-AT) \cite{madry2018towards} and TRADES \cite{zhang2019theoretically}. PGD-AT formulates AT as a min-max optimization problem and trains a model with only adversarial examples. TRADES minimizes a regularized surrogate loss to obtain a better trade-off between robustness and performance, where both clean data and adversarial examples are used for training. However, AT requires the labels of input images to generate the corresponding adversarial examples, which can not be directly applied to the UDA problem. In this work, we propose a new AT method specifically designed for UDA.
	
	%-------------------------------------------------------------------------
	\section{Preliminary}
	\noindent \textbf{UDA.}
	Given a labeled source dataset $\mathbb{D}_s = \{(x_s^i, y_s^i)\}_{i=1}^{n_s}$ and an unlabeled target dataset $\mathbb{D}_t = \{x_t^i\}_{i=1}^{n_t}$ with $n_{s}$ and $n_{t}$ number of samples, respectively, a typical UDA model learns a feature extractor $F$ and a classifier $C$ on top of $F$. Given an input image $x$, we express its feature space representation as $F(x)$ and its output logits as $C(x)$, where we use $C(x)$ as a simplification of the formal expression $C(F(x))$. The objective function of an UDA model can be written as:
	% UDA loss
	\begin{equation}
	\mathcal{L}_{CE}\big(C(x_s), y_s\big) + \mathcal{L}_{DA}\big(x_s, x_t\big),
	\end{equation}
	where $\mathcal{L}_{CE}$ is the standard cross-entropy loss, and $\mathcal{L}_{DA}$ is the domain adaptation loss defined by each UDA approach. One of the most common $\mathcal{L}_{DA}$ is the adversarial loss introduced by DANN \cite{ganin2016domain}, which is defined as:
	% DANN loss
	\begin{equation}
	\mathcal{L}_{DA}\big(x_s, x_t\big) = \mathbb{E}[logD(F(x_s))] + \mathbb{E}[1 - (logD(F(x_t)))],
	\end{equation}
	where $D$ is a domain discriminator used to encourage domain-invariant features.
	
	\noindent \textbf{AT.}
	PGD-AT \cite{madry2018towards} is one of the most commonly-used AT algorithm. It formulates AT as a min-max optimization problem and trains models on adversarial examples exclusively:
	% PGD-AT
	\begin{equation} \label{eq:at}
	\mathop{\min}\limits_{F, C} \mathbb{E} \left[\mathop{\max}\limits_{\|\delta\|_p \le \epsilon} \mathcal{L}\big(C(\tilde{x}), y\big)\right],
	\end{equation}
	where $\tilde{x} = x + \delta$ is the generated adversarial example of $x$, and $\epsilon$ denotes an $L_p$-norm perturbation budget. Here, $ \delta$ corresponds to the adversarial noise which is added to $x$ to make it adversarial. For image classification tasks, $\mathcal{L}$ is the cross-entropy loss $\mathcal{L}_{CE}$. PGD \cite{madry2018towards} is the most widely-used attack approaches. It generates $\tilde{x}$ in an iterative way:
	% FGSM
	\begin{equation} \label{eq:fgsm}
	x^{j+1} = \Pi_{\|\delta\|_p \le \epsilon}\left(x^j  +\alpha \cdot sign(\bigtriangledown_{x^j} \mathcal{L}(C(x^j), y))\right);
	\end{equation}
	and $\tilde{x} = x^{j_{max}}$, where $j_{max}$ denotes the maximum number of attack iterations. FGSM \cite{goodfellow2015explaining} can be treated as a single-step and non-random start version of PGD.
	
	%-------------------------------------------------------------------------
	\section{Exploring AT for UDA} \label{sec:sec4}
	In this section, we systematically study multiple variants of AT to explore suitable AT methods for UDA. Then we finalize the proposed ARTUDA accordingly. Here we conduct a set of experiments on the VisDA-2017 \cite{peng2017visda} dataset. We employ DANN \cite{ganin2016domain} as the UDA algorithm with ResNet-50 \cite{he2016deep} backbone. The white-box FGSM \cite{goodfellow2015explaining} attack with perturbation budget of $\epsilon = 3$ is used for both AT and testing. Following the practice of \cite{awais2021adversarial,yang2021exploring}, we assume that attackers have the labels of the target dataset to generate adversarial examples. The rationale behind these settings is that (i) most existing UDA approaches \cite{long2017conditional,tzeng2017adversarial} are based on DANN's key idea, so DANN is a fair representative; (ii) the white-box threat model is the strongest attack setting, which has been considered a standard evaluation protocol for defenses \cite{madry2018towards,zhang2019theoretically,obfuscated,xie2019intriguing}. 
	
	%-------------------------------------------------------------------------
	\subsection{Conventional AT on UDA}
	We start with applying a conventional AT \cite{madry2018towards} to DANN to see its effectiveness. That is, the AT is performed on only the labeled source data, \textit{i.e.}, apply Eq.~(\ref{eq:at}) on source dataset $\mathbb{D}_s$. Therefore, the objective of the DANN model becomes:
	% UDA loss with AT
	\begin{equation}
	\mathcal{L}_{CE}\big(C(\tilde{x}_s), y_s\big) + \mathcal{L}_{DA}\big(\tilde{x}_s, x_t\big). 
	\end{equation}
	It is reasonable to expect that Conventional AT cannot fully benefit target domain robustness, as source domain robustness may not perfectly transfer to the target domain due to domain shift. As reported in Table~\ref{table:unsupervised_at}, compared to the Natural Training baseline (\textit{i.e.}, train with only clean data), Conventional AT indeed improves robustness to a certain extent but is not significant. Also, the clean accuracy is largely decreased. Hence, we argue that applying AT directly on the target data is important.
	
	A naive way of applying AT on the target data is to produce pseudo labels $y'_t$ using an external pre-trained UDA model. ASSUDA \cite{yang2021exploring} resorts to this idea and applies it to the UDA semantic segmentation problem. Note that ASSUDA only evaluates black-box robustness. Here we implement the \textit{Pseudo Labeling} idea on image classification and observe its white-box robustness. We use a naturally trained DANN as the pseudo labeler. The objective of Pseudo Labeling approach is as follows:
	% UDA loss with Pseudo Labeling
	\begin{equation}
	\mathcal{L}_{CE}\big(C(x_s), y_s\big) + \mathcal{L}_{CE}\big(C(\tilde{x}_t), y'_t\big) +  \mathcal{L}_{DA}\big(x_s, \tilde{x}_t\big). 
	\end{equation}
	In Table~\ref{table:unsupervised_at}, we find that Pseudo Labeling's robustness is not better than Conventional AT, and the clean accuracy drops dramatically. We believe that the label noise problem is inevitable in pseudo labels $y'_t$ and limits model performance. This motivates us to explore a new AT method that can be directly performed on the target domain.
	
	% Conventional AT and Self-Supervised AT
	%\renewcommand{\arraystretch}{1.1}
	\setlength{\tabcolsep}{20pt}
	\begin{table}[!t]
		\begin{center}
			\caption{Results (\%) of Conventional AT and our Self-Supervised AT on the VisDA-2017 dataset.}
			\label{table:unsupervised_at}
			\begin{tabular}{l | c | c}
				\hline \noalign{\smallskip} \noalign{\smallskip}
				Training method & Clean & FGSM \\
				\noalign{\smallskip} \hline \noalign{\smallskip}
				Natural Training & 73.2 & 21.2 \\
				Conventional AT \cite{madry2018towards} & 62.9 (-10.3) & 27.1 (+5.9) \\
				Pseudo Labeling & 33.1 (-40.1) & 27.1 (+5.9) \\
				\noalign{\smallskip} \hline \noalign{\smallskip}
				Self-Supervised AT-L1 & 56.2 (-17.0) & 15.8 (-5.4) \\
				Self-Supervised AT-L2 & 51.3 (-21.9) & 26.0 (+4.8) \\
				Self-Supervised AT-KL & 67.1 \textbf{(-6.1)} & \textbf{35.0 (+13.8)} \\
				\noalign{\smallskip} \hline
			\end{tabular}
		\end{center}
		%\vspace*{-\baselineskip}
	\end{table}
	
	\subsection{Self-Supervised AT}
	Inspired by \cite{kannan2018adversarial,zhang2019theoretically}, we seek to use clean target data's logits $C(x_t)$ as a self-supervision signal to generate adversarial examples $\tilde{x}_t$. Based on the min-max optimization for AT \cite{madry2018towards}, we generate $\tilde{x}_t$ by maximizing the difference between $C(x_t)$ and $C(\tilde{x}_t)$, and minimize that difference to train a model. With this idea, we can generate adversarial examples via self-supervision and perform a kind of AT for the target domain. We call it \textit{Self-Supervised Adversarial Training} or \textit{Self-Supervised AT}. In other words, to generate $\tilde{x}_t$, Self-Supervised AT changes Eq.~(\ref{eq:fgsm}) to:
	% Self-Supervised FGSM
	\begin{equation}
	x_t^{j+1} = \Pi_{\|\delta\|_p \le \epsilon}\left(x_t^j  +\alpha \cdot sign(\bigtriangledown_{x_t^j} \mathcal{L}(C(x_t^j), C(x_t)))\right),
	\end{equation}
	and $\tilde{x}_t = x_t^{j_{max}}$. To adversarially train an UDA model, Self-Supervised AT changes Eq.~(\ref{eq:at}) to:
	% Self-Supervised AT
	\begin{equation}
	\mathop{\min}\limits_{F, C} \mathbb{E} \left[\mathop{\max}\limits_{\|\delta\|_p \le \epsilon} \mathcal{L}\big(C(\tilde{x}_t), C(x_t)\big)\right].
	\end{equation}
	$\mathcal{L}$ is a loss function that encourages the logits to be similar. Possible choices include L1 loss, L2 loss, Kullback-Leibler (KL) divergence loss, \textit{etc}. Taking KL divergence loss as an example, the objective of Self-Supervised AT for UDA can be written as follows:
	% Self-Supervised AT for UDA
	\begin{equation}
	\mathcal{L}_{CE}\big(C(x_s), y_s\big) + \mathcal{L}_{KL}\big(C(\tilde{x}_t), C([x_t]_{sg})\big) +  \mathcal{L}_{DA}\big(x_s, \tilde{x}_t\big),
	\end{equation}
	where $[\cdot]_{sg}$ denotes the stop-gradient operator \cite{van2017neural} constraining its operand to be a non-updated constant. We do not expect that Self-Supervised AT is as robust as conventional supervised AT since the ground-truth labels $y$ are always the strongest supervision. However, given that target domain labels $y_t$ are unavailable, we believe that the clean logits $C(x_t)$ could be a good self-supervision signal.
	
	Table~\ref{table:unsupervised_at} shows that Self-Supervised AT-L1 and Self-Supervised AT-L2 are not effective, while Self-Supervised AT-KL achieves excellent results. Self-Supervised AT-KL increases robust accuracy over Natural Training by 13.8\%, which is much better than Conventional AT. It also maintains decent clean accuracy. These results demonstrate that our Self-Supervised AT strategy is effective, but the choice of the loss function is critical, where KL divergence loss is the preferred one.
	
	% Combining conventional AT and SS-AT
	\setlength{\tabcolsep}{2.5pt}
	\begin{table}[!t]
		\begin{center}
			\caption{Results (\%) of SS-AT variants on VisDA-2017. $(x_s, x_t)$ denotes $\mathcal{L}_{DA}(x_s, x_t)$. $\bullet$: selected. ---: not applicable.}
			\label{table:combination}
			\begin{tabular}{l | cccc | cccc | cc}
				\hline \noalign{\smallskip} \noalign{\smallskip}
				Training method & $x_s$ & $\tilde{x}_s$ & $x_t$ & $\tilde{x}_t$ & $(x_s, x_t)$ & $(x_s, \tilde{x}_t)$ & $(\tilde{x}_s, x_t)$ & $(\tilde{x}_s, \tilde{x}_t)$ & Clean & FGSM \\
				\noalign{\smallskip} \hline \noalign{\smallskip}
				Natural Training & $\bullet$ &  & $\bullet$ &  & $\bullet$ & ----- & ----- & ----- & 73.2 & 21.2 \\
				Conventional AT \cite{madry2018towards} &  & $\bullet$ & $\bullet$ &  & ----- & ----- & $\bullet$ & ----- & 62.9 & 27.1 \\
				SS-AT-KL & $\bullet$ &  &  & $\bullet$ & ----- & $\bullet$ & ----- & ----- & 67.1 & 35.0 \\
				\noalign{\smallskip} \hline \noalign{\smallskip}
				SS-AT-s-t-\~{t}-1 & $\bullet$ &  & $\bullet$ & $\bullet$ & $\bullet$ &  & ----- & ----- & 67.3 & 27.5 \\
				SS-AT-s-t-\~{t}-2 & $\bullet$ &  & $\bullet$ & $\bullet$ & $\bullet$ & $\bullet$ & ----- & ----- & 73.0 & 39.4 \\
				SS-AT-s-\~{s}-t-\~{t}-1 & $\bullet$ & $\bullet$ & $\bullet$ & $\bullet$ & $\bullet$ &  &  & $\bullet$ & 63.4 & 41.6 \\
				SS-AT-s-\~{s}-t-\~{t}-2 &$\bullet$  & $\bullet$ & $\bullet$ & $\bullet$ &  & $\bullet$ & $\bullet$ &  & 62.8 & 42.3 \\
				SS-AT-s-\~{s}-t-\~{t}-3 & $\bullet$ & $\bullet$ & $\bullet$ & $\bullet$ & $\bullet$ & $\bullet$ & $\bullet$ & $\bullet$ & 61.3 & 41.6 \\
				\noalign{\smallskip} \hline
			\end{tabular}
		\end{center}
		%\vspace*{-\baselineskip}
	\end{table}
	
	\subsection{On the Effects of Clean and Adversarial Examples in Self-Supervised AT.} \label{sec:43}
	Let us revisit the results of the last experiment from another perspective. We observe a trade-off between clean performance and robustness, and the upper part of Table~\ref{table:combination} illustrates this point more clearly. Specifically, from Natural Training and Conventional AT, we can see that replacing clean images $x_s$ by adversarial examples $\tilde{x}_s$ increases robust accuracy but decreases clean accuracy. A similar trade-off can be found between Natural Training and Self-Supervised AT-KL, which train with $x_t$ and $\tilde{x}_t$, respectively. This interests us to further investigate the usage of the four data types $\{x_s, \tilde{x}_s, x_t, \tilde{x}_t\}$ in the AT for UDA. Self-Supervised AT-KL outperforms Conventional AT in terms of both clean and robust accuracies, indicating that using $\tilde{x}_t$ is more efficient than $\tilde{x}_s$, so we start with Self-Supervised AT-KL as a baseline.
	
	First, we add $x_t$ to Self-Supervised AT-KL. This turn out SSAT-s-t-\~{t}-1 and SSAT-s-t-\~{t}-2, where SSAT-s-t-\~{t}-1's domain adaptation loss is $\mathcal{L}_{DA}(x_s, x_t)$, while SSAT-s-t-\~{t}-2 involves another term and becomes $\mathcal{L}_{DA}(x_s, x_t) + \mathcal{L}_{DA}(x_s, \tilde{x}_t)$. In other words, SSAT-s-t-\~{t}-1 explicitly transfers the supervised knowledge from $x_s$ to only $x_t$, while SSAT-s-t-\~{t}-2 transfers to both $x_t$ and $\tilde{x}_t$. We expect that SSAT-s-t-\~{t}-1 and SSAT-s-t-\~{t}-2 enjoy higher clean accuracy than Self-Supervised AT-KL because they involve $x_t$.
	
	The lower part of Table~\ref{table:combination} reports the results. We find that SSAT-s-t-\~{t}-1's robust accuracy drops significantly, but the clean accuracy does not improve much. In contrast, SSAT-s-t-\~{t}-2 largely increases both clean and robust accuracies by 5.9\% and 4.4\%, respectively. The improvement of clean performance matches our expectations, but we are surprised at that of robustness. We see this is due to our Self-Supervised AT's specific property. Self-Supervised AT leverages the objective $\mathcal{L}_{KL}\big(C(\tilde{x}_t), C(x_t)\big)$ to do AT, so $C(x_t)$'s quality is critical. Given that the labels $y_t$ is unavailable, $\mathcal{L}_{DA}(x_s, x_t)$ can transfer the supervised knowledge to $x_t$ and thus enhance $C(x_t)$'s quality. Therefore, adding $x_t$ to Self-Supervised AT benefits robustness as well. This observation is different from the conventional supervised AT that exists the trade-off between performance and robustness \cite{tsipras2018robustness,xie2019intriguing,zhang2019theoretically}. We conclude that involving $x_t$ into training does help, but an explicit supervised knowledge transfer to $\tilde{x}_t$ is needed. This is rational since $\tilde{x}_t$ plays the most important role in Self-Supervised AT, giving firm guidance to it is essential.
	
	Second, we look into the effects of $\tilde{x}_s$ in Self-Supervised AT. We add $\tilde{x}_s$ and study three variants: SSAT-s-\~{s}-t-\~{t}-1, SSAT-s-\~{s}-t-\~{t}-2 and SSAT-s-\~{s}-t-\~{t}-3. Their differences are in their domain adaptation loss, which is also illustrated in Table~\ref{table:combination}. Intuitively, we expect that adding $\tilde{x}_s$ falls into the trade-off that leads to lower clean performance but better robustness, as $\tilde{x}_s$ is the conventional supervised adversarial example.
	
	As shown in Table~\ref{table:combination}, all the three variants obtain lower clean accuracy and higher robust accuracy than SSAT-s-t-\~{t}-1 and SSAT-s-t-\~{t}-2, which matches our assumption. The results among these three are very close. Compared to SSAT-s-t-\~{t}-2, their clean accuracy drops 9.6\%-11.7\%, but robust accuracy only improves 2.2\%-2.9\%. This is consistent with Conventional AT's result, \textit{i.e.}, source domain robustness is not easy to transfer to the target domain. Because training without $\tilde{x}_s$ achieves a better trade-off between performance and robustness, we use SSAT-s-t-\~{t}-2 as a baseline for the next investigation. To present our experiments more clear, in the following, we summarize the objective functions of each Self-Supervised AT variant discussed in this part:
	
	\noindent -- SSAT-s-t-\~{t}-1:
	\begin{equation}
	\mathcal{L}_{CE}\big(C(x_s), y_s\big) + \mathcal{L}_{KL}\big(C(\tilde{x}_t), C([x_t]_{sg})\big) + \mathcal{L}_{DA}\big(x_s, x_t\big).
	\end{equation}
	\noindent -- SSAT-s-t-\~{t}-2:
	\begin{equation} \label{eq:artuda}
	\begin{split}
	& \mathcal{L}_{CE}\big(C(x_s), y_s\big) + \mathcal{L}_{KL}\big(C(\tilde{x}_t), C([x_t]_{sg})\big) \\ & + \mathcal{L}_{DA}\big(x_s, x_t\big) + \mathcal{L}_{DA}\big(x_s, \tilde{x}_t\big).
	\end{split} 
	\end{equation}
	\noindent -- SSAT-s-\~{s}-t-\~{t}-1:
	\begin{equation}
	\begin{split}
	& \mathcal{L}_{CE}\big(C(x_s), y_s\big) + \mathcal{L}_{KL}\big(C(\tilde{x}_t), C([x_t]_{sg})\big) \\ & + \mathcal{L}_{CE}\big(C(\tilde{x}_s), y_s\big) + \mathcal{L}_{DA}\big(x_s, x_t\big) + \mathcal{L}_{DA}\big(\tilde{x}_s, \tilde{x}_t\big).
	\end{split} 
	\end{equation}
	\noindent -- SSAT-s-\~{s}-t-\~{t}-2:
	\begin{equation}
	\begin{split}
	& \mathcal{L}_{CE}\big(C(x_s), y_s\big) + \mathcal{L}_{KL}\big(C(\tilde{x}_t), C([x_t]_{sg})\big) \\ & + \mathcal{L}_{CE}\big(C(\tilde{x}_s), y_s\big) + \mathcal{L}_{DA}\big(x_s, \tilde{x}_t\big) + \mathcal{L}_{DA}\big(\tilde{x}_s, x_t\big).
	\end{split} 
	\end{equation}
	\noindent -- SSAT-s-s-'t-\~{t}-3:
	\begin{equation}
	\begin{split}
	& \mathcal{L}_{CE}\big(C(x_s), y_s\big) + \mathcal{L}_{KL}\big(C(\tilde{x}_t), C([x_t]_{sg})\big) + \mathcal{L}_{CE}\big(C(\tilde{x}_s), y_s\big) \\ & + \mathcal{L}_{DA}\big(x_s, x_t\big) + \mathcal{L}_{DA}\big(x_s, \tilde{x}_t\big) + \mathcal{L}_{DA}\big(\tilde{x}_s, x_t\big) + \mathcal{L}_{DA}\big(\tilde{x}_s, \tilde{x}_t\big).
	\end{split} 
	\end{equation}
	
	\subsection{On the Effects of BN in Self-Supervised AT}
	It has been well-known that the statistic estimation of BN \cite{ioffe2015batch} plays an important role in both the UDA \cite{chang2019domain,li2017revisiting} and the adversarial machine learning \cite{lo2021defending,xie2019intriguing,xie2020adversarial} fields. It is worth investigating the effects of BN given these two research fields meet together in this paper.
	
	Recall that during training, BN computes the mean and variance of the feature space for each mini-batch, referred to as \textit{batch statistics} \cite{xie2019intriguing}. Each mini-batch is normalized by its batch statistics at training time. Hence, the composition of a mini-batch defines its batch statistics, thereby affecting the normalized values of each data point's features. To observe the effects on Self-Supervised AT, we create four variants of SSAT-s-t-\~{t}-2. They involve the same data types $\{x_s, x_t, \tilde{x}_t\}$ into training but with different mini-batch compositions. Specifically, at each training step, Batch-st-\~{t} has two mini-batches, [$x_s$, $x_t$] and [$\tilde{x}_t$]; Batch-s-t\~{t} has two mini-batches, [$x_s$] and [$x_t$, $\tilde{x}_t$]; Batch-s-t-\~{t} has three mini-batches, [$x_s$], [$x_t$] and [$\tilde{x}_t$]; and Batch-st\~{t} has one mini-batch, [$x_s$, $x_t$, $\tilde{x}_t$]. Batch-st-\~{t} is the original SSAT-s-t-\~{t}-2, which follows the setting of \cite{dalib}. We expect that their batch statistics differences would cause different results.
	
	Table~\ref{table:batchnorm} shows the results. As can be seen, Batch-st-\~{t} achieves the highest clean accuracy, while Batch-st\~{t} achieves the highest robust accuracy. We argue that in Batch-st\~{t}, $x_s$ is with the same mini-batch as $x_t$ and $\tilde{x}_t$, so it can also transfer the supervised knowledge through batch statistics. In other words, the batch statistics used to normalize $x_t$ and $\tilde{x}_t$ contain $x_s$'s information. This shares a similar spirit with the domain adaptation loss $\mathcal{L}_{DA}\big(x_s, x_t\big) + \mathcal{L}_{DA}\big(x_s, \tilde{x}_t\big)$ discussed in Sec.~\ref{sec:43}, and we have known that it can improve robustness. For Batch-st-\~{t}, we see its high performance is due to the separation of $x_t$ and $\tilde{x}_t$. Recall that clean and robust features have distinct characteristics \cite{itazuri2019adversarially,tsipras2018robustness}, so putting them into the same mini-batch leads to suboptimal results \cite{xie2019intriguing}. Batch-s-t-\~{t}, however, achieves lower performance than Batch-st-\~{t} though it has that separation as well. The reason is that in Batch-st-\~{t}, $x_s$ and $x_t$ are with the same mini-batch. This encourages the knowledge transfer from $x_s$ to $x_t$, similar to the spirit of the domain adaptation loss $\mathcal{L}_{DA}\big(x_s, x_t\big)$.
	
	Both Batch-st-\~{t} and Batch-st\~{t} achieve a good trade-off between performance and robustness. We can choose according to the downstream application's focus.
	
	% Batch combination
	\setlength{\tabcolsep}{20pt}
	\begin{table}[!t]
		\begin{center}
			\caption{Results (\%) of different mini-batch combinations on the VisDA-2017 dataset.}
			\label{table:batchnorm}
			\begin{tabular}{l | c | cc}
				\hline \noalign{\smallskip} \noalign{\smallskip}
				Method & Mini-batches & Clean & FGSM \\
				\noalign{\smallskip} \hline \noalign{\smallskip}
				Batch-st-\~{t} & [$x_s$, $x_t$], [$\tilde{x}_t$] & 73.0 & 39.4 \\
				Batch-s-t\~{t} & [$x_s$], [$x_t$, $\tilde{x}_t$] & 68.2 & 37.0 \\
				Batch-s-t-\~{t} & [$x_s$], [$x_t$], [$\tilde{x}_t$] & 68.2 & 35.5 \\
				Batch-st\~{t} & [$x_s$, $x_t$, $\tilde{x}_t$] & 69.0 & 41.4 \\
				\noalign{\smallskip} \hline
			\end{tabular}
		\end{center}
		%\vspace*{-\baselineskip}
	\end{table}
	
	\subsection{Summary}
	In this section, we explore four main aspects of AT for UDA, including Conventional AT, our Self-Supervised AT, the effects of clean and adversarial examples in Self-Supervised AT, and the effects of BN statistics. We progressively derive the best method from each investigation, then we take Batch-st\~{t} as our final method, referred to as Adversarially Robust Training for UDA (ARTUDA). ARTUDA's training objective is Eq.(\ref{eq:artuda}), and Fig.~\ref{fig:main} offers a visualized illustration. Note that some of the other variants also have their advantages, e.g., Batch-st-\~{t}, so they are still useful for certain focusses.
	
	%-------------------------------------------------------------------------
	\section{Experiments} \label{sec:expts}
	We extensively evaluate the proposed ARTUDA on five adversarial attacks, three datasets and three different UDA algorithms. We further compare ARTUDA with the nearly contemporary work, RFA \cite{awais2021adversarial}. An analysis of feature space is also presented.
	
	%-------------------------------------------------------------------------
	\subsection{Experimental Setup}
	
	\noindent \textbf{Datasets.}
	We use three UDA datasets for evaluation: VisDA-2017 \cite{peng2017visda}, Office-31 \cite{saenko2010adapting} and Office-Home \cite{venkateswara2017deep}. VisDA-2017 contains two domains: Synthetic and Real. There are 152,409 Synthetic and 55,400 Real images from 12 object categories in this large-scale dataset. Office-31 has three domains with 31 object categories. These are Amazon (A) with 2,817 images, Webcam (W) with 795 images, and DSLR (D) with 498 images. We employ the D $\to$ W task for our expeiment. Office-Home includes four domains with 65 categories: Art (Ar) with 2,427 images, Clipart (Cl) with 4,365, Product (Pr) with 4,439 images, and Real-World (Rw) with 4,375 images. We employ the Ar $\to$ Cl task for our experiment.
	
	\noindent \textbf{Attack setting.}
	We test UDA models' adversarial robustness against four white-box attacks, including FGSM \cite{goodfellow2015explaining}, PGD \cite{madry2018towards}, MI-FGSM \cite{dong2018boosting} and MultAdv \cite{lo2021multav}, where PGD is the default attack unless stated otherwise. A black-box attack \cite{papernot2017practical} is also considered. For AT, we use the PGD attack with $j_{max}=3$ and $\epsilon=3$ of $L_\infty$-norm. If not otherwise specified, we set the same for all the attacks at testing time except that FGSM's $j_{max}$ is $1$.
	
	\noindent \textbf{Benchmark UDA algorithms.}
	We apply ARTUDA to three common UDA algorithms, including DANN \cite{ganin2016domain}, JAN \cite{long2017deep} and CDAN \cite{long2017conditional}. We use ResNet-50 \cite{he2016deep} as a backbone for all of them. If not otherwise specified, DANN is the default UDA algorithm in our experiments.
	
	\noindent \textbf{Baseline defenses.}
	We employ two commonly-used conventional AT algorithms, PGD-AT \cite{madry2018towards} and TRADES \cite{zhang2019theoretically}, to be our baseline defenses. To the best of our knowledge, RFA \cite{awais2021adversarial} might be the only approach aimming at the same problem as ours, and we also compare with it.
	
	\noindent \textbf{Implementation details.}
	Our implementation is based on PyTorch \cite{paszke2019pytorch}. We adopt Transfer-Learning-library \cite{dalib} to set up UDA's experimental environment and follow the training hyper-parameters used in \cite{dalib}. We also use the widely-used library, AdverTorch \cite{ding2019advertorch}, to perform adversarial attacks. We will release our source code if the paper gets accepted.
	
	%-------------------------------------------------------------------------
	\subsection{Evaluation Results} \label{sec:results}
	
	\noindent \textbf{White-box robustness.}
	The robustness of multiple training methods against various white-box attacks is reported in Table~\ref{table:main_results_1}. Without a defense, Natural Training's accuracy drops to almost 0\% under the strong iterative attacks. PGD-AT and TRADES improve adversarial robustness though they are originally designed for the traditional classification task. However, they also reduce clean accuracy. The proposed method, ARTUDA, significantly increases robust accuracy. Specifically, on VisDA-2017, it achieves more than 10\% and 20\% higher robustness than TRADES and PGD-AT, respectively. On Office-31, its robust accuracy is higher than PGD-AT and TRADES by 25\%-48\% under white-box iterative attacks. On Office-Home, although TRADES is slightly more robust to white-box iterative attacks, ARTUDA has higher accuracy under clean data, FGSM and black-box attacks, leading by a decent margin. In general, ARTUDA is effective across all the five attacks on three datasets. ARTUDA's clean accuracy drops but is still the best among the defenses. It can greatly improve robustness and maintain decent clean performance simultaneously.
	
	\noindent \textbf{Black-box robustness.}
	The robustness against black-box attacks is shown in the last column of Table~\ref{table:main_results_1}. Here we consider a naturally trained DANN with ResNet-18 as a substitute model and use MI-FGSM, which has better transferability, to generate black-box adversarial examples for target models. In general, the black-box attacks hardly fool the target models. However, we find that the conventional AT approaches have lower black-box accuracy than Natural Training in some cases. This is due to their lower clean accuracy. In contrast, ARTUDA has better clean accuracy and consistently achieves the best black-box robustness across all the datasets.
	
	% Main results 1
	\setlength{\tabcolsep}{4.5pt}
	\begin{table}[!t]
		\scriptsize
		\begin{center}
			\caption{Results (\%) of UDA models on multiple datasets under various adversarial attacks.}
			\label{table:main_results_1}
			\begin{tabular}{l | l | c | cccc | c}
				\hline \noalign{\smallskip} \noalign{\smallskip}
				Dataset & Training method & Clean & FGSM & PGD & MI-FGSM & MultAdv & Black-box \\
				\noalign{\smallskip} \hline \noalign{\smallskip}
				& Natural Training & 73.2 & 21.2 & 0.9 & 0.5 & 0.3 & 58.3 \\
				VisDA-2017 & PGD-AT \cite{madry2018towards} & 60.5 & 34.6 & 21.3 & 22.7 & 7.8 & 59.1 \\
				\cite{peng2017visda} & TRADES \cite{zhang2019theoretically} & 64.0 & 42.1 & 29.7 & 31.2 & 16.4 & 62.6 \\
				& ARTUDA (ours) & 65.5 & \textbf{52.5} & \textbf{44.3} & \textbf{45.0} & \textbf{27.3} & \textbf{65.1} \\
				\noalign{\smallskip} \hline \noalign{\smallskip}
				& Natural Training & 98.0 & 52.7 & 0.9 & 0.6 & 0.1 & 95.0 \\
				Office-31 & PGD-AT \cite{madry2018towards} & 95.3 & 91.8 & 68.2 & 66.5 & 31.4 & 95.3 \\
				D $\to$ W\cite{saenko2010adapting} & TRADES \cite{zhang2019theoretically} & 88.4 & 85.3 & 66.4 & 67.0 & 28.2 & 88.2 \\
				& ARTUDA (ours) & 96.5 & \textbf{95.2} & \textbf{92.5} & \textbf{92.5} & \textbf{77.1} & \textbf{96.5} \\
				\noalign{\smallskip} \hline \noalign{\smallskip}
				& Natural Training & 54.5 & 26.4 & 4.7 & 2.8 & 2.0 & 53.1 \\
				Office-Home & PGD-AT \cite{madry2018towards} & 42.5 & 38.8 & 36.0 & 35.8 & 21.7 & 43.0 \\
				Ar $\to$ Cl \cite{venkateswara2017deep} & TRADES \cite{zhang2019theoretically} & 49.3 & 45.1 & \textbf{41.6} & \textbf{41.6} & \textbf{22.5} & 49.4 \\
				& ARTUDA (ours) & 54.0 & \textbf{49.}5 & 41.3 & 39.9 & 21.6 & \textbf{53.9} \\
				\noalign{\smallskip} \hline
			\end{tabular}
		\end{center}
		%\vspace*{-\baselineskip}
	\end{table}
	
	% Main results 2
	\setlength{\tabcolsep}{2pt}
	\begin{table}[!t]
		\scriptsize
		\begin{center}
			\caption{Results (\%) of UDA models on the VisDA-2017 dataset under the PGD attack. Three UDA algorithms are considered.}
			\label{table:main_results_2}
			\begin{tabular}{l | ccc | ccc | ccc}
				\hline \noalign{\smallskip} \noalign{\smallskip}
				UDA algorithm $\to$ &  &  \underline{\, DANN \cite{ganin2016domain} \,} &  &  & \underline{\, JAN \cite{long2017deep} \,} &  &  & \underline{\, CDAN \cite{long2017conditional} \,} & \\
				Training method $\downarrow$ & Clean & PGD & Drop & Clean & PGD & Drop & Clean & PGD & Drop \\
				\noalign{\smallskip} \hline \noalign{\smallskip}
				Natural Training & 73.2 & 0.0 & -73.2 & 64.2 & 0.0 & -64.2 & 75.1 & 0.0 & -75.1 \\
				PGD-AT \cite{madry2018towards} & 60.5 & 13.3 & -47.2 & 47.7 & 5.8 & -41.9 & 58.2 & 11.7 & -46.5 \\
				TRADES \cite{zhang2019theoretically} & 64.0 & 19.4 & -44.6 & 48.7 & 8.5 & -40.2 & 64.6 & 15.7 & -48.9 \\
				Robust PT \cite{awais2021adversarial} & 65.8 & 38.2 & -27.6 & 55.1 & 32.2 & \textbf{-22.9} & 68.0 & 41.7 & -26.3 \\
				RFA \cite{awais2021adversarial} & 65.3 & 34.1 & -31.2 & 63.0 & 32.8 & -30.2 & 72.0 & 43.5 & -28.5 \\
				\noalign{\smallskip} \hline \noalign{\smallskip}
				ARTUDA (ours) & 65.5 & \textbf{40.7} & \textbf{-24.8} & 58.5 & \textbf{34.4} & -24.1 & 68.0 & \textbf{43.6} & \textbf{-24.4} \\
				\noalign{\smallskip} \hline
			\end{tabular}
		\end{center}
		%\vspace*{-\baselineskip}
	\end{table}
	
	\noindent \textbf{Generalizability.}
	To compare with the results of \cite{awais2021adversarial}, in this part, we evaluate robustness against the white-box PGD attack with $j_{max}=20$ that used in \cite{awais2021adversarial}. Table~\ref{table:main_results_2} reports the adversarial robustness of multiple popular UDA algorithms. All of them are vulnerable to adversarial attacks. The state-of-the-art approaches, Robust PT and RFA, show excellent effectiveness in improving robustness. We apply our ARTUDA training method to these UDA models to protect them as well. As can be seen, ARTUDA uniformly robustfies all of these models. It consistently achieves low accuracy drops and the highest robust accuracy, which outperforms both Robust PT and RFA. This demonstrates that ARTUDA is generic and can be applied to multiple existing UDA algorithms.
	
	In terms of clean data accuracy, all the defenses lose clean accuracy to a certain extent. Still, the proposed ARTUDA achieves the best or the second-best clean accuracy among these defenses. Overall, it can significantly improve robustness and maintain decent clean performance simultaneously.
	
	% L2 distance
	\begin{figure}[!t]
		\centering
		\includegraphics[width=0.7\textwidth]{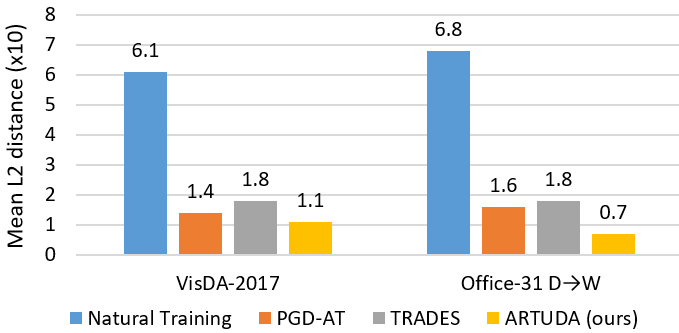}
		\caption{Mean $L_2$-norm distance between the feature space of clean images and that of their adversarial examples. The values are the mean over an entire dataset.}
		\label{fig:distance}
	\end{figure}
	
	% tSNE
	\begin{figure}[!t]
		\centering
		\includegraphics[width=0.8\textwidth]{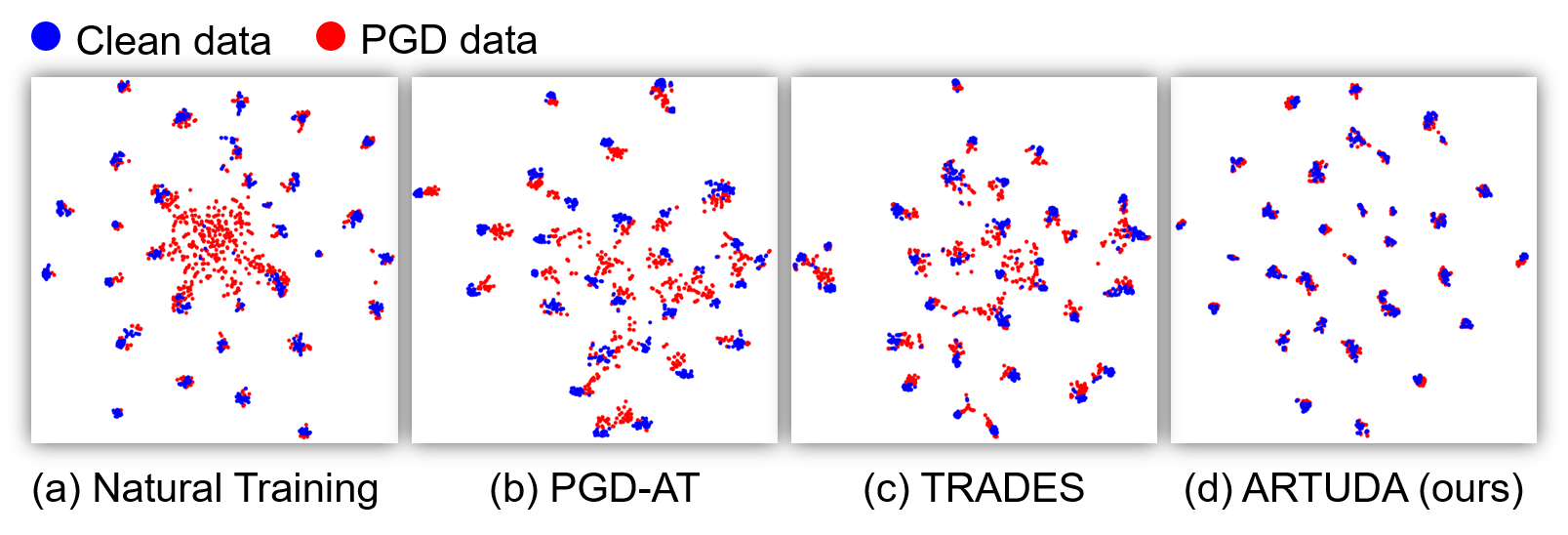}
		\caption{The t-SNE visualization of the feature space on the Office-31 D$\to$W task.}
		\label{fig:tsne}
		%\vspace*{-\baselineskip}
	\end{figure}
	
	%-------------------------------------------------------------------------
	\subsection{Analysis}
	\noindent \textbf{Stability of feature space.}
	Small adversarial perturbations on image space are enlarged considerably in feature space \cite{Xie_2019_CVPR}. Hence, the stability of the feature space can reflect a model's robustness \cite{lo2021error}. In other words, a robust model's feature space would hardly change under an adversarial example. We compute the mean $L_2$-norm distance between the feature space of clean images and that of their PGD examples for our models: $\parallel F(x_t) - F(\tilde{x}_t) \parallel_2$. The features from the last conv layer of the ResNet-50 backbone are used. As can be seen in Fig.~\ref{fig:distance}, Natual Training has the largest distance, which means that its features are greatly changed when images are adversarially perturbed and thus cause wrong predictions. PGD-AT and TRADES can reduce the distance. ARTUDA attains the smallest distance on both datasets, showing that its feature space is not easily affected by adversarial perturbations.
	
	\noindent \textbf{Visualization of feature space.}
	Fig.~\ref{fig:tsne} visualizes the different methods' feature space on the Office-31 D$\to$W task using t-SNE \cite{van2008visualizing}. The features are from the last conv layer of the ResNet-50 backbone. The PGD data in the Natural Training model are disorderly scatter and do not align with clean data. PGD-AT and TRADES narrow the distribution gap to a certain extent. ARTUDA impressively align the feature space of PGD and clean data in which they almost overlap with each other. This implies that ARTUDA is effective in learning adversarially robust features. This result is consistent with the above stability analysis.
	
	% Attack budget
	\begin{figure}[!t]
		\centering
		\includegraphics[width=0.75\textwidth]{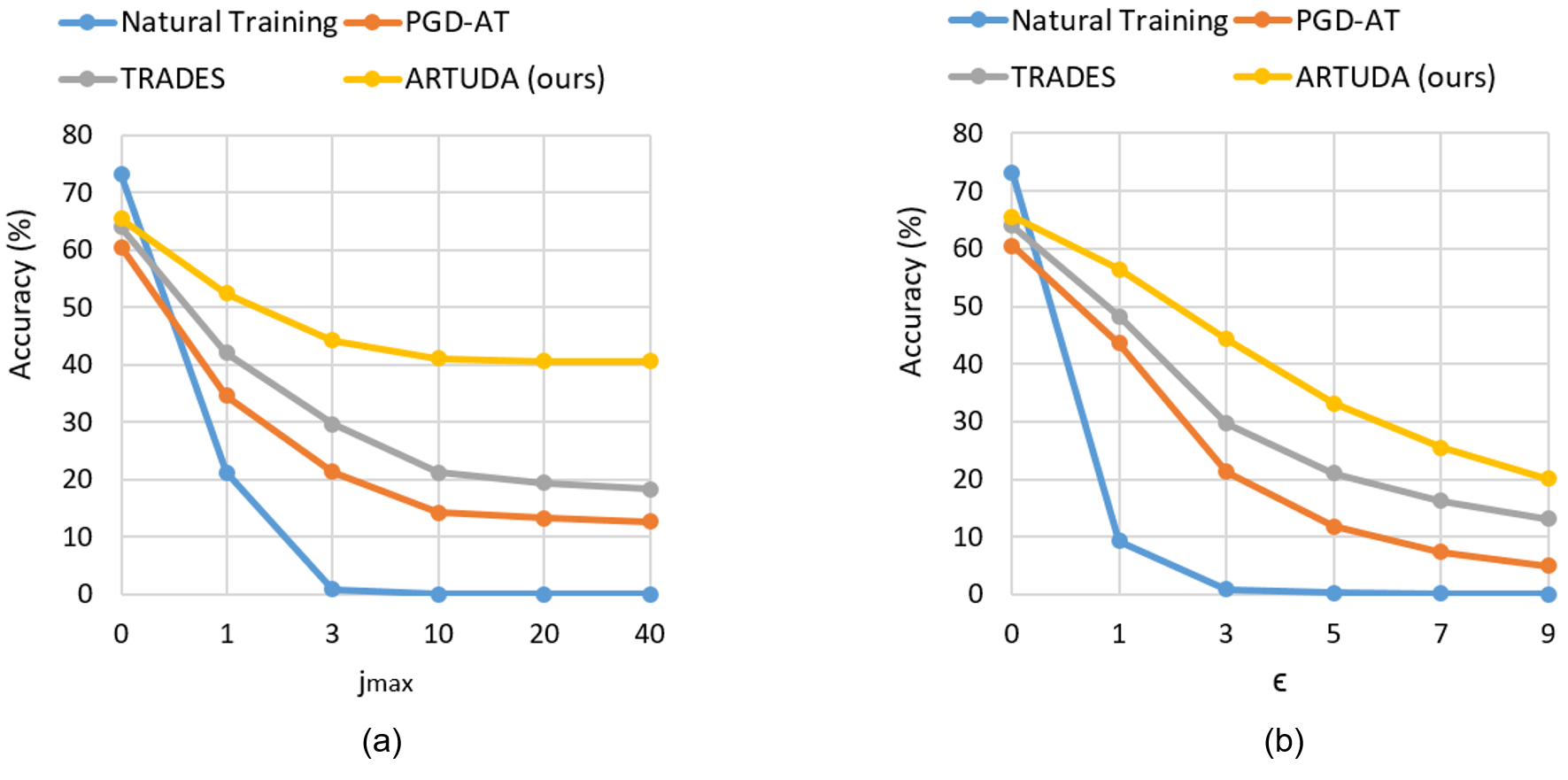}
		\caption{Accuracy of models under PGD attacks (a) with varied numbers of attack iterations $j_{max}$ and (b) with varied perturbation sizes $\epsilon$.}
		\label{fig:budget}
		%\vspace*{-\baselineskip}
	\end{figure}
	
	\noindent \textbf{Attack budgets.}
	We test our ARTUDA's scalability to various attack budgets. We vary the attack budgets by two aspects: the number of attack iterations $j_{max}$ and the perturbation size $\epsilon$. Fig.~\ref{fig:budget} shows the results. First, we can find that the attack strength does not increase apparently along with the increase of $j_{max}$ when $j_{max} > 3$. This observation is consistent with that of \cite{madry2018towards}. The proposed ARTUDA demonstrates stable adversarial robustness and consistently performs better than Natural Training, PGD-AT \cite{madry2018towards} and TRADES \cite{zhang2019theoretically} under varied $j_{max}$. On the other hand, the attack strength dramatically increases along with the increase of $\epsilon$. It can be seen that ARTUDA consistently shows better robustness under varied $\epsilon$. Obviously, ARTUDA is scalable to various attack budgets.

	%------------------------------------------------------------------------
	\section{Conclusion}
	This paper explores AT methods for the UDA problem. Existing AT approaches require labels to generate adversarial examples and train models, but this does not apply to the unlabeled target domain. We provide a systematicac study into multiple AT variants that may suitable for UDA. This empirical contribution could offer useful insight to the research community. Based on our study, we propose ARTUDA, a novel AT method specifically designed for UDA. Our comprehensive experiments show that ARTUDA improves robustness consistently across multiple attacks and datasets, and outperforms the state-of-the-art methods.
	
	\subsubsection{Acknowledgements.}
	This work was supported by the DARPA GARD Program HR001119S0026-GARD-FP-052.
	
	%===========================================================
	\bibliographystyle{splncs04}
	\bibliography{mycite}
	
	%this would normally be the end of your paper, but you may also have an appendix
	%within the given limit of number of pages

	%------------------------------------------------------------------------
\clearpage
\setcounter{section}{0}
\renewcommand\thesection{A\arabic{section}}

%\section*{Supplementary Materials of Paper 818}

\section{Basic Sanity Checks to Evaluation}
To further verify the reliability of our evaluation, we report our results on the basic sanity checks introduced in Athalye \textit{et al.} \cite{obfuscated}.
\begin{itemize}
	\item Table~\ref{table:main_results_1} shows that iterative attacks (PGD \cite{madry2018towards} and MI-FGSM \cite{dong2018boosting}) are stronger than one-step attacks (FGSM \cite{goodfellow2015explaining}).
	\item Table~\ref{table:main_results_1} shows that white-box attacks are stronger than black-box attacks \cite{papernot2017practical} (by MI-FGSM).
	\item Unbounded attacks reach 100\% attack success rate (accuracy drops to 0.0\%) on all the three datasets.
	\item Fig.~\ref{fig:budget} shows that increasing distortion bound increases attack success (decreases accuracy). 
\end{itemize}

%------------------------------------------------------------------------
\section{Loss Weight of Self-Supervised AT}
We can impose a hyperparameter on our ARTUDA training scheme. Specifically, we can add a loss weight $\lambda$ to Eq.~(\ref{eq:artuda}), and it is shown as follows:
\begin{equation}
\begin{split}
& \mathcal{L}_{CE}\big(C(x_s), y_s\big) + \lambda \mathcal{L}_{KL}\big(C(\tilde{x}_t), C([x_t]_{sg})\big) \\ & + \mathcal{L}_{DA}\big(x_s, x_t\big) + \mathcal{L}_{DA}\big(x_s, \tilde{x}_t\big).
\end{split} 
\end{equation}
The loss weight $\lambda$ controls the ratio of the Self-Supervised AT objective to the overall objective. In all of our previous experiments, we set $\lambda$ to $1$. In this section, we train multiple ARTUDA models with varied $\lambda$, where we use the experimental setup described in Sec.~\ref{sec:sec4}. The results are reported in Table~\ref{table:lambda}.

We can find that the robust accuracy significantly increases along with the increase of $\lambda$, while the clean accuracy does not vary obviously. This implies that the robustness of the proposed ARTUDA can be further improved with a larger $\lambda$ though it already outperforms the state-of-the-art methods.

% Lambda
\setlength{\tabcolsep}{20pt}
\begin{table}
	\begin{center}
		\caption{Results (\%) of ARTUDA models with varied hyperparameter $\lambda$.}
		\label{table:lambda}
		\begin{tabular}{c | cc}
			\hline \noalign{\smallskip} \noalign{\smallskip}
			$\lambda$ & Clean & FGSM \\
			\noalign{\smallskip} \hline \noalign{\smallskip}
			0.2 & 68.9 & 33.3 \\
			0.5 & 66.1 & 39.3 \\
			1.0 & 69.0 & 41.1 \\
			2.0 & 66.5 & 48.5 \\
			5.0 & 68.0 & 54.4 \\
			\noalign{\smallskip} \hline
		\end{tabular}
	\end{center}
\end{table}

%------------------------------------------------------------------------
\section{Class-wise Accuracy on VisDA-2017}
In Table~\ref{table:perclass}, we report class-wise accuracy under PGD attacks \cite{madry2018towards} on the VisDA-2017 dataset \cite{peng2017visda}. The results correspond to the PGD column in Table~\ref{table:main_results_1}. We can see that ARTUDA achieves the best accuracy across the majority of the classes.

% VisDA-2017 per-class-eval
\setlength{\tabcolsep}{1.5pt}
\begin{table}
	\scriptsize
	\begin{center}
		\caption{Class-wise accuracy (\%) under PGD attacks on the VisDA-2017 dataset.}
		\label{table:perclass}
		\begin{tabular}{l | cccccccccccc | c}
			\hline \noalign{\smallskip} \noalign{\smallskip}
			Training method & aero & bicycle & bus & car & horse & knife & motor & person & plant & skate & train & truck & Mean \\
			\noalign{\smallskip} \hline \noalign{\smallskip}
			Natural Training & 4.8 & 0.9 & 1.5 & 0.0 & 0.2 & 0.9 & 0.3 & 3.2 & 0.2 & 0.1 & 0.7 & 0.0 & 0.9 \\
			PGD-AT \cite{madry2018towards} & 49.6 & 20.4 & 15.2 & 8.7 & 34.3 & 7.3 & 27.3 & 32.8 & 35.2 & \textbf{17.4} & 19.8 & 3.2 & 21.3  \\
			TRADES \cite{zhang2019theoretically} & 61.8 & 24.5 & 32.0 & 11.4 & 42.9 & 30.6 & 34.1 & \textbf{49.1} & 50.1 & 5.6 & 33.1 & 4.8 & 29.7 \\
			\noalign{\smallskip} \hline \noalign{\smallskip}
			ARTUDA (ours) & \textbf{75.0} & \textbf{32.1} & \textbf{61.5} & \textbf{25.9} & \textbf{53.3} & \textbf{65.1} & \textbf{66.4} & 48.2 & \textbf{52.3} & 9.2 & \textbf{58.8} & \textbf{7.8} & \textbf{44.3} \\
			\noalign{\smallskip} \hline
		\end{tabular}
	\end{center}
\end{table}

\end{document}